# Clustering techniques selection for a hybrid regression model: a case study based on a solar thermal system

María Teresa García-Ordás[a], Héctor Alaiz-Moretón[a], José-Luis Casteleiro-Roca[b], Esteban Jove[b], José Alberto Benítez-Andrades[a*], Isaías García-Rodríguez[a], Héctor Quintián[b] and José Luis Calvo-Rolle[b]

[a]*Department of Electrical and Systems Engineering, Universidad de León, León, Spain;*
[b] *CTC, Department of Industrial Engineering, CITIC, University of A Coruña, Spain*

*Correspondence: jbena@unileon.es; Tel.: +34-987 29 36 28; Department of Electric, Systems and Automatics Engineering, University of León, Campus of Vegazana s/n, León, 24071, León, Spain

**María Teresa García-Ordás**, Ph.D. was born in León, Spain, in 1988. She received her degree in Computer Science from the University of León in 2010, and her Ph.D. in Intelligent Systems in 2017. She was a recipient of a special mention award for the best doctoral thesis on digital transformation by Tecnalia. Since 2019, she works as an assistant professor at the University of León. Her research interests include computer vision and deep learning. She has published several articles in impact journals and patents. She has participated in many conferences all over the world.

**Héctor Alaiz-Moretón,** Ph.D. received his degree in Computer Science, performing the final Project at Dublin Institute of Technology, in 2003. He received his PhD in Information Technologies in 2008 (University of Leon). He has worked as a lecturer since 2005 at the School of Engineering at the University of Leon. His research interests include knowledge engineering, machine and deep learning, networks communication, and security. He has several works published in international conferences, as well as books, more than 80 scientific publications between JCR papers, Lecture Notes and Scientific Workshops. He has been a member of scientific committees in conferences. He has headed several PhD Thesis and research competitive projects. Actually, he is the vice main of RIASC (INSTITUTE OF APPLIED SCIENCES TO CYBERSECURITY).

**José-Luis Casteleiro-Roca** received the B.S. from University of Coruña (UDC) in 2003, the M.S. in Industrial Engineering from the University of Leon in 2012, and also a Master in Maritime Engineering in 2012 in the UDC. He obtain the PhD degree in University of La Laguna in

2019. Since 2014, he has also been part of the teaching staff and researcher of the UDC as a part-time adjunct professor in the Department of Industrial Engineering. His main research interests have been centered in applying expert system technologies to the diagnosis and control systems and in intelligent systems for control engineering and optimization.

**Esteban Jove** received a M.S. degree in Industrial Engineering from the University of Leon in 2014. After two years working in the automotive industry, he joined the University of A Coruña, Spain, where he is Professor of Power Electronics in the Faculty of Engineering since 2016. He received the Ph.D in the University of La Laguna in 2020 and his research has been focused on the use of intelligent techniques for nonlinear systems modelling and anomaly detection using one-class techniques.

**José Alberto Benítez-Andrades,** Ph.D. was born in Granada, Spain, in 1988. He received his degree in Computer Science from the University of León, and a Ph.D. degree in Production and Computer Engineering in 2017 (University of Leon). He was part-time instructor who kept a parallel job from 2013 to 2018, from 2018 to 2020 he worked as a teaching assistant and from 2020 he works as an assistant professor at the University of Leon. His research interests include artificial intelligence, knowledge engineering, semantic technologies. He was a recipient of award to the Best Doctoral Thesis 2018 by Colegio Profesional de Ingenieros en Informática en Castilla y León in 2018.

**Isaías García-Rodríguez,** Ph.D. received his Bachelor degree in Industrial Technical Engineering from the University of León (Spain) in 1992 and her Master degree in Industrial Engineering from the University of Oviedo (Spain) in 1996. Isaías obtained his Ph.D. in Computer Science from the University of León in 2008, where he is currently a lecturer. His current research interests include practical applications of Software Defined Networks, Network Security and applied Knowledge Engineering techniques. He has published different scientific papers in journals, Conferences and Symposia around the world.

**Héctor Quintián,** received the B.S. degree in Industrial Engineering from the University of A Coruña in 2008, the M.S. degree in Computer Science from University of A Coruña and Ph.D. degree in Computer Science from the University of Salamanca in 2017, respectively. He is Associate Professor of Automatic Control at the Industrial Engineering Department and member of the Cybernetic Science and Technique research group, University of A Coruña, Spain. His main research areas are associated with artificial intelligence in the field of non supervised learning and reinforced learning, modeling and control engineering.

**José Luis Calvo Rolle** received the B.S. degree in Industrial Engineering from the University of A Coruña in 1998, and the M.S. and Ph.D. degrees in Industrial Engineering from the University of Leon in 2004 and 2007, respectively. He is Associate Professor of Automatic Control at the Industrial Engineering Department and the coordinator of the Cybernetic Science and Technique research group, University of A Coruña, Spain. His main research areas are associated with the development of expert system technologies for diagnosis and control systems and the application of intelligent systems for fault detection, modeling and control engineering.

# Clustering techniques selection for a hybrid regression model: a case study based on a solar thermal system

This work addresses the performance comparison between four clustering techniques with the objective of achieving strong hybrid models in supervised learning tasks. A real dataset from a bio-climatic house named Sotavento placed on experimental wind farm and located in Xermade (Lugo) in Galicia (Spain) has been collected. Authors have chosen the thermal solar generation system in order to study how works applying several cluster methods followed by a regression technique to predict the output temperature of the system.}

With the objective of defining the quality of each clustering method two possible solutions have been implemented. The first one is based on three unsupervised learning metrics (Silhouette, Calinski-Harabasz and Davies-Bouldin) while the second one, employs the most common error measurements for a regression algorithm such as Multi Layer Perceptron.

## 1. Introduction

In general terms there are a lot of different hot topics, and of course for the most of possible applications, and regardless of the field of the final use. Representative cases of them are: ecological, zero impact, environment safety, sustainability, and so on [5, 16]. Usually, these topic examples go in opposition with other issues like benefits, comfort, luxury, etc. [22, 21]. Furthermore, it is a challenge the compromise between the two trends; for instance, people like comfort homes, and therefore, it is desirable this achievement comes from renewable energies.

In relation to energy needs, renewable energies play a key role in contributing to a reduction in environmental impact and emissions [23]. Nevertheless, the impact of the power-plant implementation itself based on renewable sources has to be taken into account, there is not usually any zero impact [30].

Because it is not possible to achieve the null impact, even with the alternatives and use of renewable energies, there is a legal obligation to optimize and plan installations with maximum efficiency [36]. Moreover, the facilities performance must be measured in accordance with the right ratios and criteria with the aim of ensuring the desired minimum impact [18].

For an optimal performance of the renewable energy systems, due to some different reasons, commonly it is necessary to make predictions of the used variables for the facility right management [20]. There are many techniques to make predictions, from the traditional ones to the most advanced through the middle ones between both [4]. When a specific system to be modelled has a performance with a very non-linear component for instance, the modelling based on hybrid systems frequently gives very satisfactory results [6, 31, 11, 9, 25, 10].

When hybrid systems are used for modelling tasks, during the clustering stage frequently is used K-means method as a standard [34]. However, there are many clustering techniques with a satisfactory performance and, in a lot of cases, with a better performance versus K-means technique [34].

The present research accomplishes a performance study of two clustering techniques, Gaussian Mixture and Spectral Clustering. For comparing their be- haviour, two approaches have been implemented. Firstly a set of error non- supervised measurements and following an MLP (Multi Layer Perceptron) regressor for establishing the quality when a hybrid model is developed. The work has been accomplished over a real system based on a solar thermal panel, installed in a bioclimatic house.

The rest of the document is structured as follows. Section 2 describes briefly the case of study. After section 2, the model approach used to compare the clustering measurement is shown. After that, the techniques applied to achieve the classification are explained. Section 5 details the experiments and achieved results and finally, the conclusions and future works are exposed in Section 6.

## 2. Case study

The case study of this research is part of the installation of the Sotavento Galicia Foundation bioclimatic house. This Foundation was created with the aim of studying both new renewable energies and their use in building, and for this last point, they built the bioclimatic house.

### *Sotavento bioclimatic house*

The real house is shown in figure 1, and it was built with the aim of reducing the amount of energy consumed inside. It is located in Xermade council, in Lugo, in the Sotavento Experimental Wind Farm, that is a place where the Foundation has its own wind farm to study different types of wind turbines.

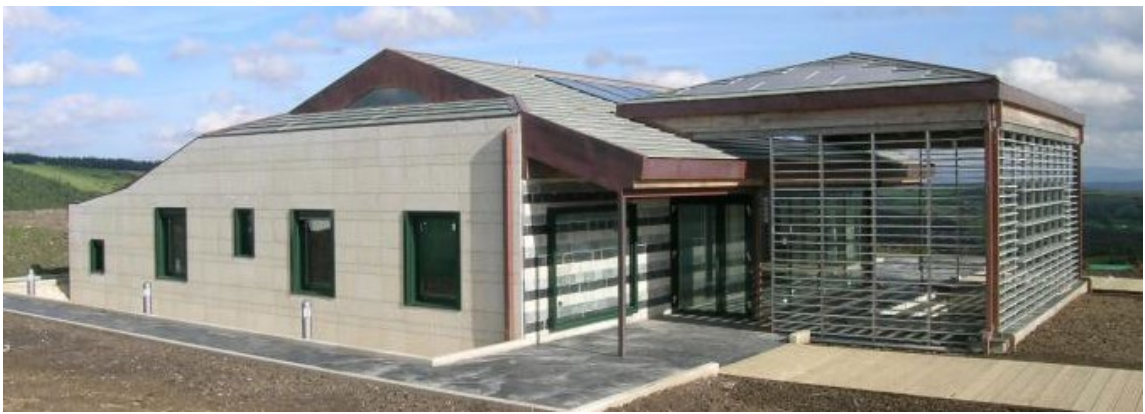

Figure 1. Sotavento bioclimatic house

This research is focused on the solar thermal energy collectors, that are only a part of the whole thermal system of the house. Figure 2 shows the thermal energy schematic of the house, which includes solar (1), biomass (2) and geothermal (3) as primary energies. The schematic is divided into three parts: generation, accumulation and consumption. The thermal energy consumption of the house is the Domestic Hot Water, DHW, (7) and the Heating system (6). The accumulation part has two different water store deposit, one is the solar accumulator (4), and the other is the DHW and Heating accumulator (5); this part also include the preheating for the DHW (8).

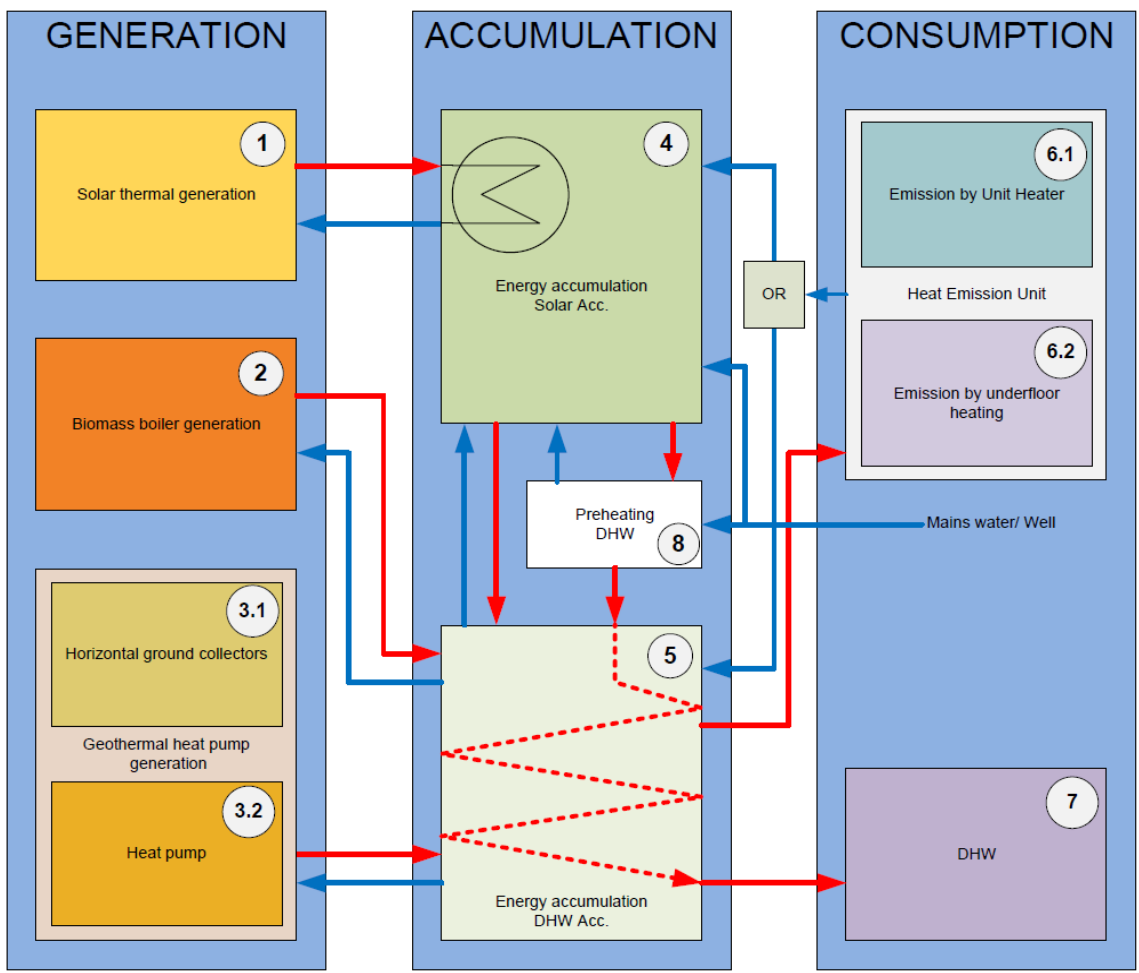

Figure 2. Thermal energy installation schematic

As the bioclimatic house is made for study not only thermal energies, the house includes also electrical energies like wind, photovoltaic and grid connection. As this type of energy is not part of the research, it is not described in this paper.

The thermal solar system is presented in figure 3, that shows the schematic of this part of the installation. This research uses only the temperature sensors S1, S2, S3 and S4, and also the flow-meter (red arrow in the figure). The solar collector (with a total surface of 20 m2) is made with eight panels, distributed in two strings of four panels. The top and the bottom string have input and output temperature sensors (S1, S2, S3 and S4); the rest of the schematic is the same for both strings. Figure 3 also includes the solar accumulator, with a capacity of 1000 L, and the necessary valves and pumps to ensure that the system could work properly.

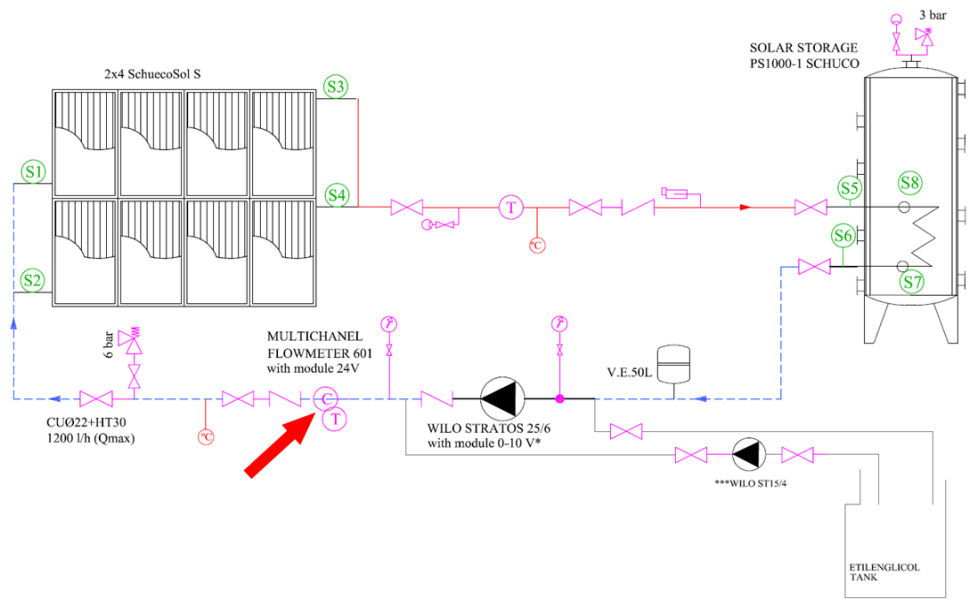

Figure 3. Solar thermal energy layout

The temperature sensors used are RTD (PT1000) type, and the flow-meter is a Multical®403. For the research we also need to know the solar radiation, that is measured with a PYR-P sensor located outside the house.

*Model approach*

The aim of this research is to compare different clustering techniques. The way used to choose the best number of clusters is based on different metrics that allow selecting to optimal number of groups. The model approach presented in this research is used to compare the clustering algorithms. Once the optimal number of clusters is chosen, a hybrid model is trained based on the schematic shown in figure 4.

Figure 4 shows the general model, but it is important to highlight that this is a hybrid model created with several local models, as many local models as clusters. The output of the model is the output temperature of the lower string (S4 in figure 3) and the inputs are:

- The inputs temperatures of both strings (S1 and S2 in figure 3).
- The flow rate of the etilenglicol used as thermal fluid throw the panels.
- The solar radiation.

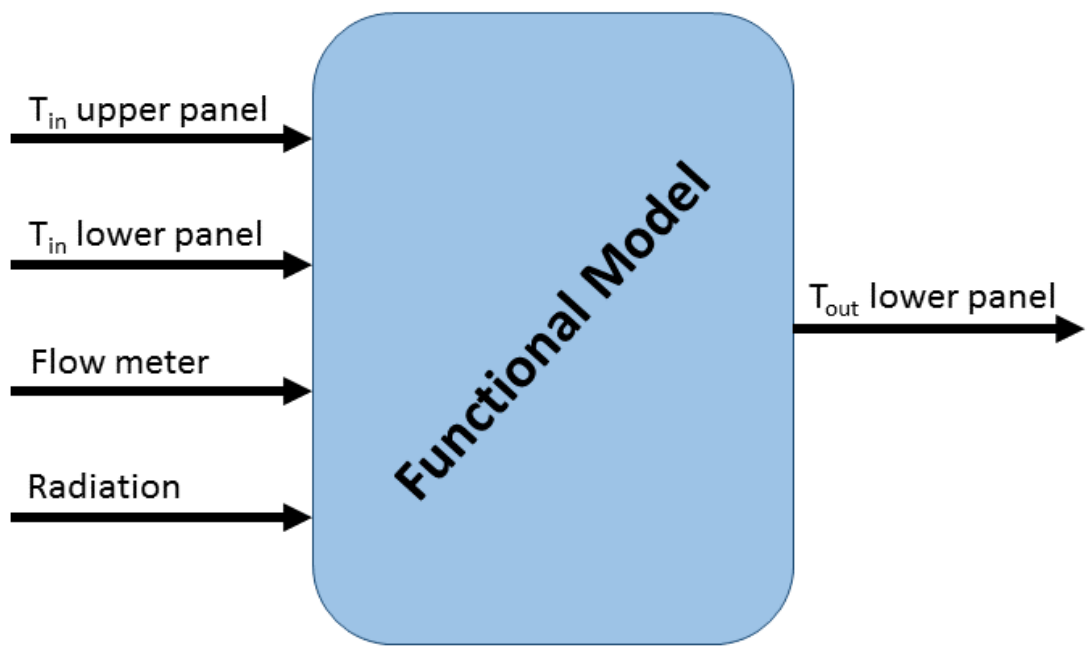

Figure 4. General schema of the functional mode

The procedure to compare the different clusters techniques, was to train one hybrid model for each technique with the same training dataset, and compare the obtained results for each one with a testing dataset. The used regression technique was an Artificial Neural Network for each internal local model.

## 3. Used techniques

The first step was a preprocessing step in which the data is normalized applying MinMax normalization. After the preprocessing step, four different clustering algorithms have been applied. Three metrics were used to evaluate each of the clustering techniques. The data class assigned by the clustering technique is used as an extra feature. After that, an MLP regressor is used to make predictions.

An LDA technique was used to improve the results visualization. A brief explanation of the implemented techniques is given in the following paragraphs.

### *3.1. Preprocessing*

The MinMax method normalizes the data to fall in the [0,1] interval depending on their maximum and minimum values, according to the following expression 1.

$$\hat{x}_j = \frac{x_j - x_{min}}{x_{max} - x_{min}} \qquad (1)$$

This normalization process is recommended to obtain better results when working with Multi-Layer Perceptron or clustering [14] techniques when used for regression analyses [3].

### *3.2. Linear Discriminant Analysis projection*

When using logistic regression, sometimes it is found that, despite classes being well separated, estimation parameters are found to be unstable. In these scenarios, the Linear Discriminant Analysis (LDA) technique is recommended because it is not affected by this kind of problem. With the aid of LDA the classes separability can be maximized.

Moreover, this technique eases data transformation, obtaining the greatest separation between classes, and so it is a good technique for projection. LDA has usually been used as a method for two-dimensional projection, as is the case in this study [27].

*3.3. Clustering techniques*

**Spectral Clustering** Spectral Clustering [28] splits a dataset according to its samples' similarity graph. Both the adjacency and the degree matrix can be obtained from this graph indicating, respectively, the relationship between samples and the number of relations. Then, the corresponding Laplacian matrix can be calculated by using the degree matrix. The final step consists in using the Laplacian matrix for applying K-means on its eigenvectors, finding the corresponding clusters of samples. Because of using K-means, the number of centroids must be previously determined.

**Gaussian Mixture Clustering** This technique [26] takes into account the centroid, the covariance and the weight for defining clusters. These models can be defined as a combination of K Gaussian distributions. An Expectation-Maximization algorithm [13] is used to find the distributions, determining the values for the mean, the covariance and the weight of each distribution. While the K-means technique only uses the mean value, the Gaussian Mixture Clustering also takes into account the variance on the data.

**Agglomerative Clustering** Agglomerative Clustering [12] is a technique included in the so-called Hierarchical Clustering family of algorithms. It builds clusters by means of a number of splitting and merging processes, starting with a unique sample per cluster. In each iteration of the algorithm, a merge between the most similar clusters is performed. The process ends when all the samples belong to the same clustering.

**K-Means algorithm** This algorithm is one of the most popular clustering techniques. K-Means tries to separate the data minimizing the inertia of the groups [17]. This method requires to set the number of clusters before training. Each cluster is represented by its centroid $\mu_j$, which represents the mean values of all its elements (see equation 2).

$$\sum_{i=0}^{n} \min_{\mu_j \in C}(||x_i - \mu_j||^2) \quad (2)$$

*3.4. Cluster Error metrics*

The unsupervised metrics Silhouette coefficient, Calinsky-Harabasz and Davies-Bouldin have been studied for evaluating the clustering methods.

**Silhouette** The Silhouette coefficient is a score for evaluating the goodness of clustering algorithms, with the objective of identifying the most adequate number of clusters.

The number of clusters when using unsupervised learning algorithms may be an input parameter or may be automatically established by the algorithm itself. When included as a parameter, as is the case with K-Means algorithm, an external score must be used to find the most adequate number of clusters. The Silhouette coefficient can be used as an indicator for estimating the ideal number of clusters, where a higher coefficient means a better quality using this number of clusters.

For an observation *j*, the Silhouette coefficient is denoted as *s(j)* and calculated as:

$$s(j) = \frac{y-x}{\max(x,y)} \quad (3)$$

Where:

- *x* is the average of distances (or dissimilarities) of observation *j* respect to the rest of observations in the cluster which *j* belongs to.

- *y* is the minimum distance to a different cluster (not the same as observation *j*). The cluster meeting this requirement is known as "the neighbourhood of *j*", and would be the second-best option for *j*.

The Silhouette score takes values between -1 and 1.

When observation *j* is on the boundary of two clusters the value of *s(j)* will be close to zero.

When *s(j)* takes a negative value, the *j* observation must be assigned to the closest cluster.

In short:

- *s(j)* ≈ 1, the assignation of the *j* observation to the cluster is correct.
- *s(j)* ≈ 0, the *j* observation lies between two different clusters.
- *s(j)* ≈ -1, the assignation of *j* observation to the cluster is wrong.

**Calinski-Harabasz** The Calinsky-Harabasz score can be obtained using the following expression (4):

$$CH = \frac{\frac{BGSS}{K-1}}{\frac{WGSS}{N-K}} = \frac{N-K}{K-1} \frac{BGSS}{WGSS} \quad (4)$$

begin *N* the number of observations and *K* the number of clusters and with

$$BGSS = \sum_{j=1}^{K} n_j ||G^j - G||^2 \quad (5)$$

(where $G^j$ denotes, for each cluster, the dispersion of the barycenters, and *G* is the barycenter of the set of data as a whole. The number of samples in the cluster $C_j$ is represented as $n_j$)

$$WGSS = \sum_{j=0}^{K} WGSS^j \quad (6)$$

$$WGSS^j = \sum_{i \in I_j} ||M_i^j - G^j||^2 \quad (7)$$

($M_i^j$ are the coefficients for the i-th row in the data matrix for cluster $C_j$, while $I_j$ represents the set of indices of the observations for the $C_j$ cluster).

**Davies-Bouldin** The Davies-Bouldin index is a score used for the evaluation of clustering algorithms. It uses characteristics and quantities that are inherent to the data set, and is defined as the mean value of the samples $M_k$ (among all the clusters), as is represented in 8.

$$DB = \frac{1}{K}\sum_{j=1}^{K} M_j \quad (8)$$

where $\delta_j$ represents the mean value distance from the points belonging to the $C_j$ cluster to their barycenter $G_j$, while $\Delta_{jj'}$ is the distance between barycenters $G^j$ and $G^{j'}$ (equation 10).

$$\Delta_{jj'} = d(G^j, G^{j'}) = ||G^j - G^{j'}|| \quad (9)$$

When the clusters are compact, smalls values are obtained for the DB index, and their corresponding centers are well separated. For this reason, the optimum number of clusters is chosen when the DB index is minimized.

$$DB = \frac{1}{K}\sum_{j=1}^{K} M_j \quad (10)$$

### 3.5. Regression Error metrics

The different regression models used in the study are compared using the fol- lowing error metrics (for all of them, the observed value is denoted by $Y_j$ and the foretold value by $\widehat{Y_j}$):

- M.A.E.: Mean Absolute Error. This metric measures differences between the real and the predicted values, having some advantages over other error scores [37].

$$MAE = \frac{1}{m}\sum_{j=1}^{m}|Y_j - \widehat{Y_j}| \quad (11)$$

- LMLS: Least Mean Log Squares. It is used as a logistic error function for both the training process and the validation error [7], equation 12.

$$LMLS = \frac{1}{m}\sum_{j=1}^{m}\log\left(1 + \frac{1}{2}(Y_j - \widehat{Y_j})^2\right) \quad (12)$$

- SMAPE: Symmetric Mean Absolute Percentage Error. The objective of this metric is to give an explanation for relative errors by using percentages [19], equation 13.

$$SMAPE = \frac{2}{m}\sum_{j=1}^{m}\frac{|Y_j - \widehat{Y_j}|}{Y_j + \widehat{Y_j}} \quad (13)$$

- MSE: Mean Squared Error. This metric can be applied in different forecast- ing problems, it can include the error variance [35] equation 14.

$$MSE = \frac{1}{m}\sum_{j=1}^{m}(Y_j - \widehat{Y_j})^2 \quad (14)$$

- MAPE: Mean Absolute Percentage Error. This metric is one of the most usual ones for measuring the accuracy of regression problems [24], equation 15.

$$MAPE = \frac{100\%}{m}\sum_{j=1}^{m}\frac{|Y_j - \widehat{Y_j}|}{Y_j} \quad (15)$$

- NMSE: Normalised Mean Square Error. This metric estimates the overall deviation between predicted and observed values [29], equation 16.

$$NMSE = \frac{1}{m}\sum_{j=1}^{m}\frac{(Y_j - \widehat{Y_j})^2}{mean(\widehat{Y_j})*mean(Y_j)} \quad (16)$$

*3.6. Regression method*

**Multi-Layer Perceptron**: A Multi-layer Perceptron (MCP) was implemented to obtain a metric for the evaluation of the previously mentioned clustering algorithms.

MLP is one of the most commonly used supervised learning techniques. The learning function for this algorithm is: $Fun(\cdot): X^n \to X^0$. The Scikit-Learn library for Python was used to implement this technique.

A cross validation procedure was used for obtaining the optimal number of neurons for the hidden layer and the best activation function for each one. With the aid of this procedure, the MLP was trained with different parameters (number of neurons and activation function) to obtain the most suitable regression model [33, 15, 8, 2].

**4. Experiments and results**

This section addresses the results from clustering and regression point of view. The first one makes reference to how clustering methods have working based on a set of measurements. On the other hand, the second one, defines how a regression technique as MLP works with the clustering procedure applied previously.

*4.1. Cluster*

Four different clustering techniques have been evaluated with the aim of determining possible groupings of the unsupervised data. These techniques are: Spectral Clustering, Gaussian Mixture Clustering, Agglomerative clustering and K-Means. After the clustering step, the assigned group of each sample is used as an extra feature for a supervised regression. A hyperparameters study was carried out varying the number of clusters and finally, to determine which is the best configuration for the presented problem, three different unsupervised metrics were taken into account: Silhouette,

Calinski-Harabasz and Davies-Bouldin scores. In table 1 we can see the results achieved with the selected hyperparameter.

| Clustering | Best number of clusters | Silhouette | Calinski-Harabasz | Davies-Bouldin |
|---|---|---|---|---|
| Gaussian Mixture | 4 | 0.4450 | 32735.4139 | 0.7654 |
| Spectral Clustering | 3 | 0.4936 | 40391.5038 | 0.6354 |
| Agglomerative Clustering | 4 | 0.5279 | 41354.7560 | 0.6359 |
| K-Means | 4 | **0.5374** | **47787.0924** | **0.6338** |

Table 1. Best hyperparameter scoring using for the clustering techniques implemented

In order to get a projected visualization of the data, a 2D mapping was done by training a LDA model using the cluster assigned to each sample as its class. In figures 5-8, we can see the 2D projection for all clustering techniques evaluated.

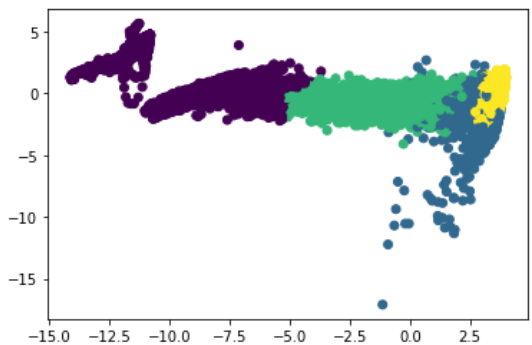

Figure 5. 2D representation of the dataset for Gaussian Mixture technique.

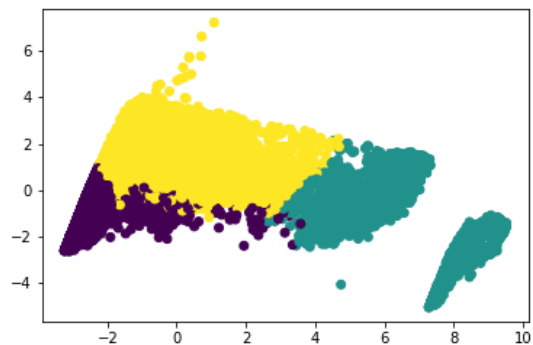

Figure 6. 2D representation of the dataset for Spectral clustering technique.

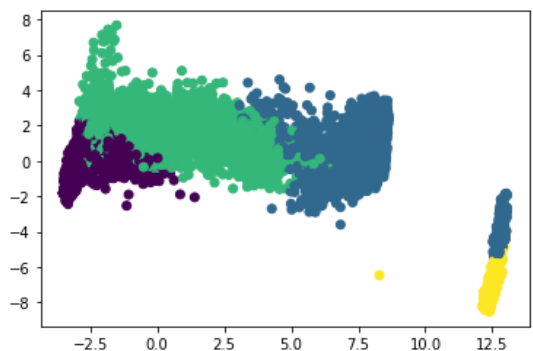

Figure 7. 2D representation of the dataset for Agglomerative clustering technique.

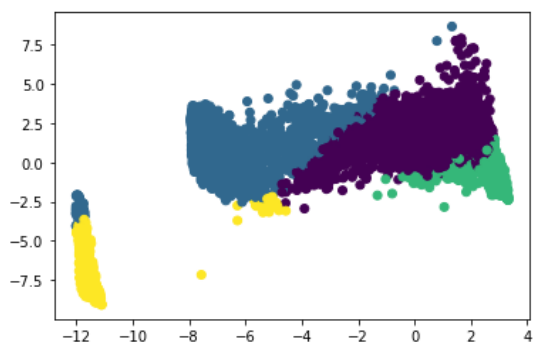

Figure 8. 2D representation of the dataset for K-Means clustering technique.

*4.2. Regression*

The main challenge of this work is to find the best clustering method for developing hybrid models, therefore it is necessary to complement the clustering metrics, due to this kind of metrics gives an idea about how the cluster processes is working, from unsupervised learning point of view. For this reason, in order to achieve efficient hybrid models applying clustering techniques, is essential to join them with a regression method. In this case MLP architecture is used.

An optimal behavior of MLP is based in the correct election of parameters for each cluster extracted by each clustering method applied. With this purpose, a Grid Search join to Cross-validation procedure has been designed in order to get the best parameters [1], with Mean Squared Error [32] like measure for defining the optimal model in the training process. The combination of parameters tested are showing following:

- Number of neurons in the hidden layer: from 12 to 30 neurons.
- Activation function: the hyperbolic tan function ("tanh") or the rectified linear unit function ("relu").
- Solver: optimizer in the family of quasi-Newton methods ("lbfgs"), stochastic gradient descent ("sgd") or stochastic gradient-based optimizer ("adam").

Final results show the four different approaches that have been implemented, based on the four clustering methods addressed previously. The tables 2, 3, 4 and 5 reflect each error measurement based on a weighted average, proportional for each error measure to the size of each grouping. The validation split is formed by the 20% of the total number of cases (5333).

| Cluster | 1 | 2 | 3 | 4 | Weighted average |
|---|---|---|---|---|---|
| MSE | 24.444 | 0.576 | 60.793 | 15.529 | **28.898** |
| MAE | 3.074 | 0.563 | 6.014 | 2.793 | **3.321** |
| LMLS | 1.386 | 0.201 | 2.468 | 1.334 | **1.423** |
| MAPE | 0.283 | 0.010 | 0.269 | 0.124 | **0.218** |
| MASE | 0.184 | 0.036 | 0.597 | 0.280 | **0.262** |
| SMAPE | 0.112 | 0.010 | 0.234 | 0.125 | **0.123** |

Table 2. MLP error for Gaussian Mixture clustering with 4 clusters

| Cluster | 1 | 2 | 3 | Weighted average |
|---|---|---|---|---|
| MSE | 27.356 | 2.358 | 57.282 | **35.635** |
| MAE | 3.283 | 0.891 | 5.870 | **3.948** |
| LMLS | 1.474 | 0.373 | 2.440 | **1.677** |
| MAPE | 0.080 | 0.025 | 0.323 | **0.176** |
| MASE | 0.277 | 0.046 | 0.743 | **0.441** |
| SMAPE | 0.079 | 0.025 | 0.261 | **0.151** |

Table 3. MLP error for Spectral clustering with 3 clusters

| Cluster | 1 | 2 | 3 | 4 | Weighted average |
|---|---|---|---|---|---|
| MSE | 56.655 | 1.191 | 60.793 | 11.128 | **27.447** |
| MAE | 5.8624 | 0.699 | 2.897 | 2.578 | **3.331** |
| LMLS | 2.437 | 0.279 | 1.317 | 1.295 | **1.446** |
| MAPE | 0.324 | 0.013 | 0.070 | 0.135 | **0.134** |
| MASE | 0.692 | 0.049 | 0.252 | 0.479 | **0.346** |
| SMAPE | 0.254 | 0.013 | 0.070 | 0.125 | **0.113** |

Table 4. MLP error for Agglomerative clustering with 4 clusters

| Cluster | 1 | 2 | 3 | 4 | Weighted average |
|---|---|---|---|---|---|
| MSE | 5.2092 | 59.4698 | 21.2747 | 16.5433 | **36.7701** |
| MAE | 0.8098 | 6.0651 | 2.9002 | 3.0794 | **4.0445** |
| LMLS | 0.305 | 2.5043 | 1.321 | 1.492 | **1.704** |
| MAPE | 0.014 | 0.326 | 0.069 | 0.136 | **0.187** |
| MASE | 0.056 | 0.752 | 0.277 | 0.4 | **0.470** |
| SMAPE | 0.014 | 0.263 | 0.068 | 0.137 | **0.155** |

Table 5. MLP error for K-Means clustering with 4 clusters

Tables 6 shows the best parameters for each MLP model extracted from the list implemented on Grid Search procedure.

Figures 9, 10, 11 and 12 shows the graphical representation for each group where MLR regressor was applied, being the "Y" axis the output value, which refers to the output temperature of the lower solar panel situated in the output system. Each graphic displays the predicted output represented in red and the real output represented in blue. Only 100 elements from each data sample have been displayed for visualization purposes, when the size of the cluster is large enough, due to there are several clusters than contain minus than 100 data cases.

| **Gaussian clustering** | | | | |
|---|---|---|---|---|
| **Grid Parameter / Cluster** | **1** | **2** | **3** | **4** |
| **Number of neurons** | 25 | 23 | 30 | 30 |
| **Activation function** | tanh | tanh | tanh | tanh |
| **Solver** | lbfgs | lbfgs | lbfgs | lbfgs |
| **Spectral clustering** | | | | |
| **Grid Parameter / Cluster** | **1** | **2** | **3** | |
| **Number of neurons** | 27 | 27 | 27 | |
| **Activation function** | tanh | tanh | tanh | |
| **Solver** | lbfgs | lbfgs | lbfgs | |
| **Agglomerative clustering** | | | | |
| **Grid Parameter / Cluster** | **1** | **2** | **3** | **4** |

| Number of neurons | 30 | 30 | 29 | 18 |
|---|---|---|---|---|
| Activation function | tanh | tanh | tanh | tanh |
| Solver | lbfgs | lbfgs | lbfgs | lbfgs |
| **K-Means clustering** | | | | |
| Grid Parameter / Cluster | **1** | **2** | **3** | **4** |
| Number of neurons | 21 | 27 | 24 | 25 |
| Activation function | tanh | tanh | tanh | tanh |
| Solver | lbfgs | lbfgs | lbfgs | lbfgs |

Table 6. MLP best parameters for each clustering algorithm

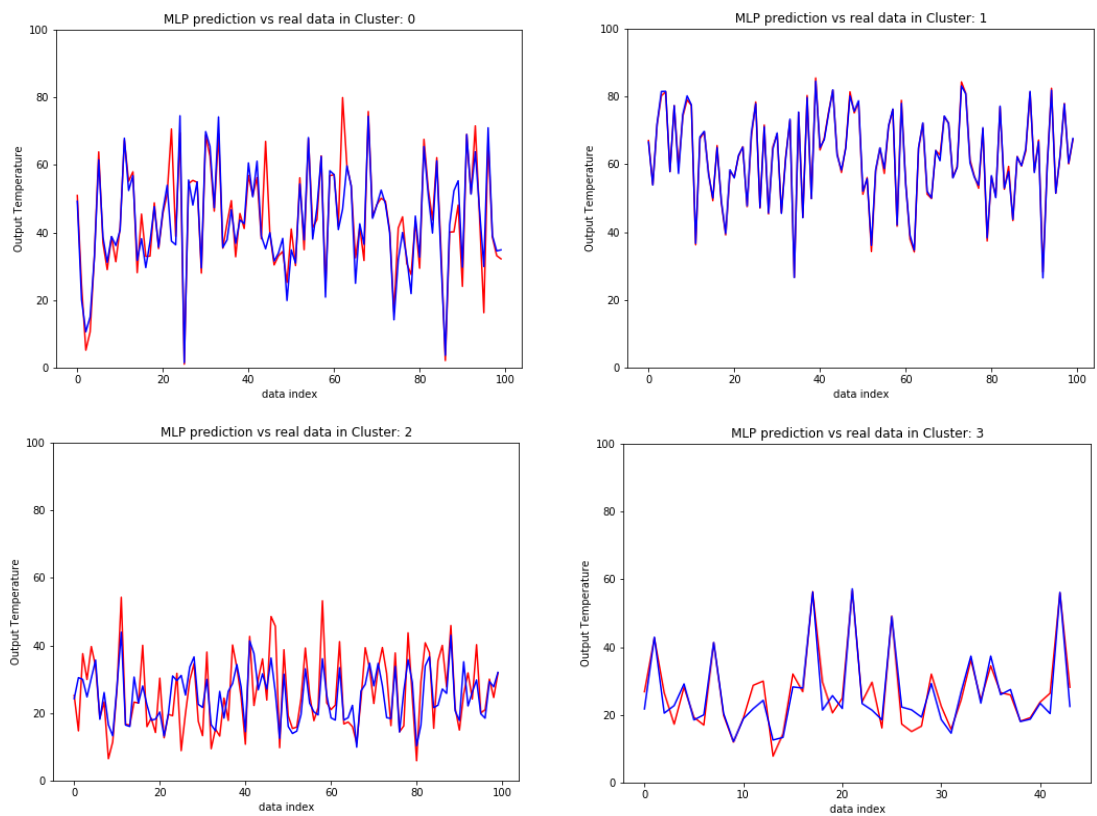

Figure 9. Real data vs. MLP predictions for Gaussian Mixture clustering

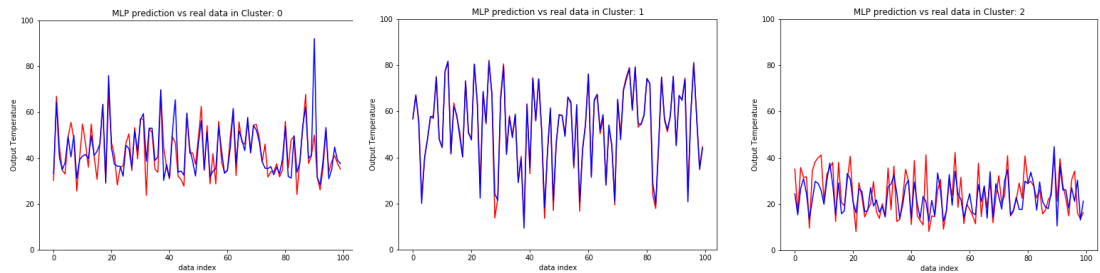

Figure 10. Real data vs. MLP predictions for Spectral clustering

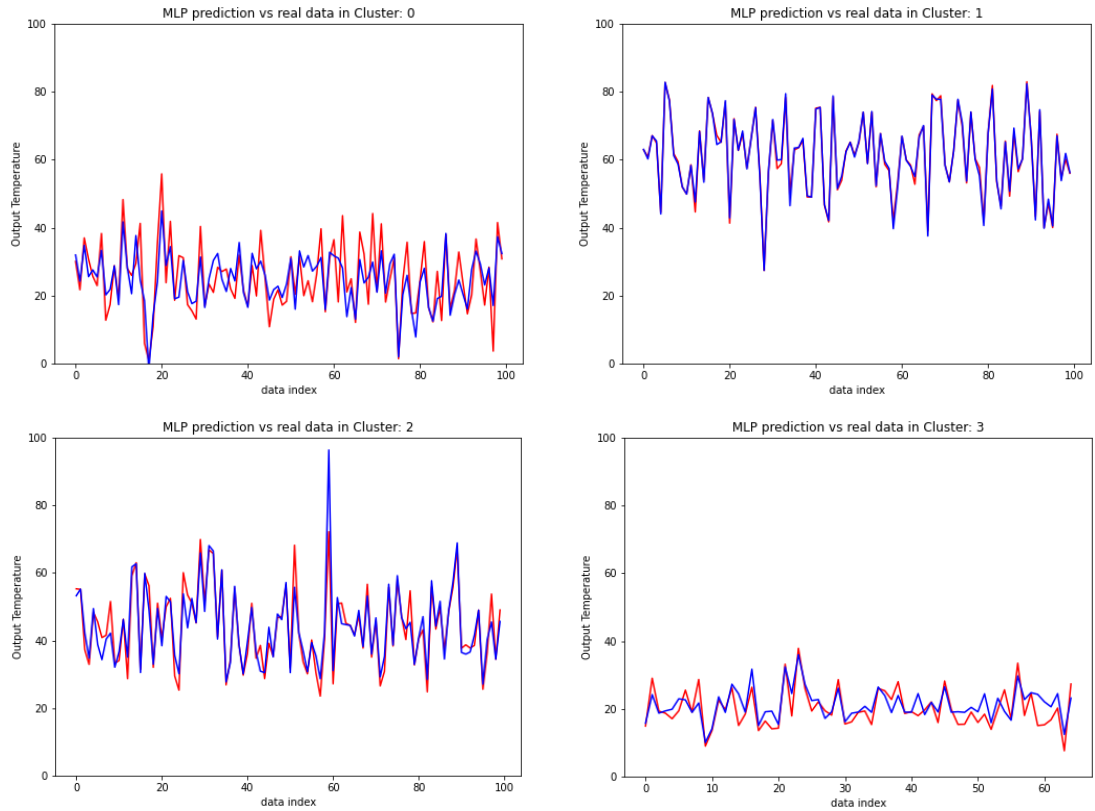

Figure 11. Real data vs. MLP predictions for Gaussian Mixture clustering

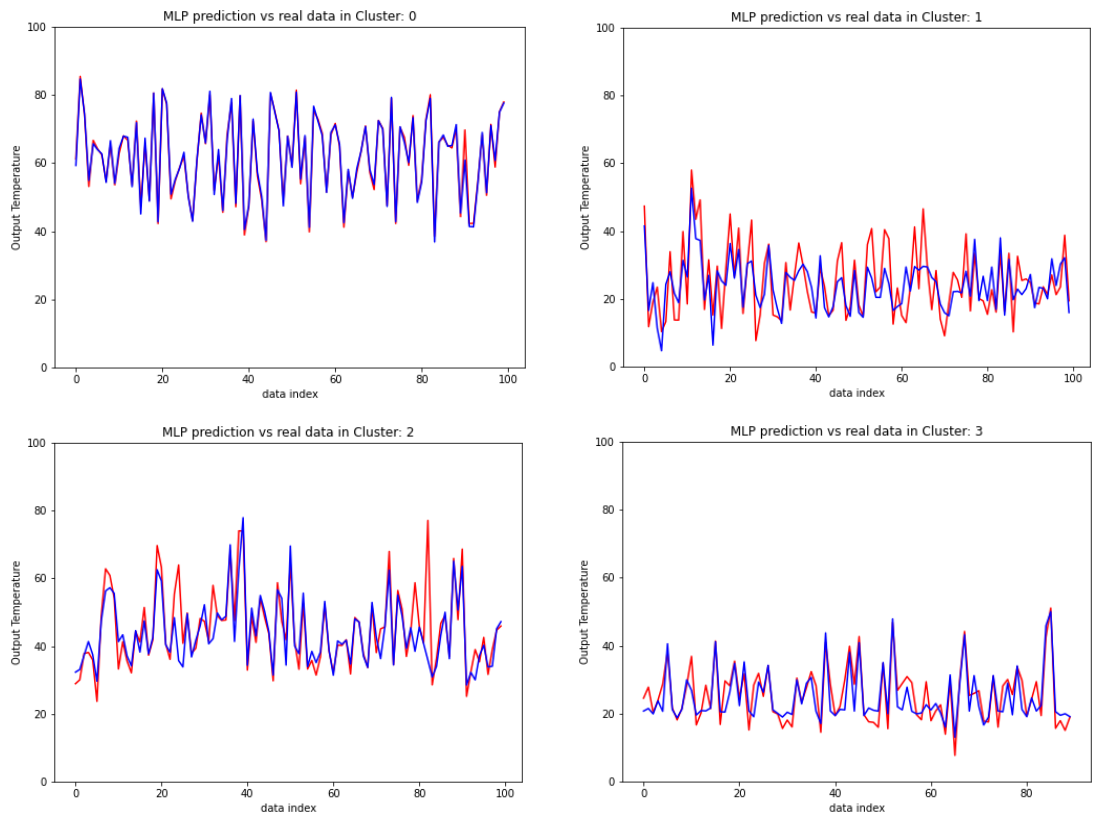

Figure 12. Real data vs. MLP predictions for Gaussian Mixture clustering

*4.3 Discussion*

Showing the unsupervised clustering results, it can be concluded that in all cases except Spectral Clustering the best number of clusters is 4. K-Means algorithm shows the best performance for the three evaluated metrics. Although all the results are quite similar, the worst results are achieved by Gaussian Mixture with the worst values for all the metrics. Agglomerative clustering and Spectral Clustering shown a similar clustering results with a slightly better values in first one taking into account that Spectral Clustering just groups the data in 3 clusters while Agglomerative Clustering groups it in 4.

As we can see in tables 2-5, the results achieved on the MLP show two different groups of performance. The better results are achieved with Gaussian Mixture and Agglomerative clustering with a similar errors in all the metrics evaluated. However, Agglomerative clustering improves Gaussian mixture in a 5% taking into account the Mean Squared Error. In the other point, both spectral clustering and K-Means show a lower performance than the other two clustering methods being outperformed by Agglomerative in more than a 25%.

Taking into account the results of the clustering with Shilouette, Calinski-Harabasz and Davies-Bouldin, agglomerative clustering has demonstrated to obtain good clustering power and a good regression power in combination with MLP. In contrast, K-Means achieved a great performance in clustering but showing high errors in regression.

**5. Conclusions and future works**

The paper address four possible clustering methods: Gaussian Mixture Clustering, Spectral Clustering, Agglomerative Clustering and K-means, in order to achieve the best one for implementing a robust hybrid MLP regression model of a thermal solar system. Based on the typical regression error metrics and specific clustering errors metrics such as Silhouette, Calinski-Harabasz and Davies-Bouldin, authors can conclude that the best method is Agglomerative Clustering, being the optimal number of cluster four.

While it is true that four groups could limit the effective operations of MLP, the combination of this technique with Agglomerative Clustering could even be better working with bigger datasets.

Future works will be oriented to apply other regression techniques such as Support Vector Machines, Extra Tree Regressor and Polynomial Regression. On the other hand, authors will work with new real datasets from bio-climatic field.